%% file: main.tex
\documentclass[10pt,twocolumn,letterpaper]{article}

\usepackage{cvpr}              
\usepackage{multirow}

\definecolor{cvprblue}{rgb}{0.21,0.49,0.74}
\usepackage[pagebackref,breaklinks,colorlinks,allcolors=cvprblue]{hyperref}

\def\paperID{14287} 
\def\confName{CVPR}
\def\confYear{2025}

\title{Style Evolving along Chain-of-Thought for Unknown-Domain Object Detection}

\author{Zihao Zhang\textsuperscript{1}, Aming Wu\textsuperscript{2}, Yahong Han\textsuperscript{1}\thanks{Corresponding author.}\\
\textsuperscript{1}College of Intelligence and Computing, Tianjin University, Tianjin, China\\
\textsuperscript{2}School of Electronic Engineering, Xidian University, Xi’an, China\\
{\tt\small zhangzihao2490@tju.edu.cn, amwu@xidian.edu.cn,  yahong@tju.edu.cn}
}

\usepackage[ruled,vlined,linesnumbered]{algorithm2e}
\usepackage{colortbl}
\definecolor{customcolor}{HTML}{dbeef3} 

\begin{document}
\maketitle
\input{sec/0_abstract}    
\input{sec/1_intro}
\input{sec/2_relatework}

\input{sec/3_method}
\input{sec/4_experiment}
\input{sec/5-Conclusion}

{
    \small
    \bibliographystyle{ieeenat_fullname}
    \bibliography{main}
}


\end{document}

%% file: sec/0_abstract.tex
\begin{abstract}
Recently, a task of Single-Domain Generalized Object Detection (Single-DGOD) is proposed,  aiming to generalize a detector to multiple unknown domains never seen before during training. Due to the unavailability of target-domain data, some methods leverage the multimodal capabilities of vision-language models, using textual prompts to estimate cross-domain information, enhancing the model's generalization capability. These methods typically use a single textual prompt, often referred to as the one-step prompt method. However, when dealing with complex styles such as the combination of rain and night, we observe that the performance of the one-step prompt method tends to be relatively weak. The reason may be that many scenes incorporate not just a single style but a combination of multiple styles. The one-step prompt method may not effectively synthesize combined information involving various styles. To address this limitation, we propose a new method, i.e., Style Evolving along Chain-of-Thought, which aims to progressively integrate and expand style information along the chain of thought, enabling the continual evolution of styles. Specifically, by progressively refining style descriptions and guiding the diverse evolution of styles, this approach enables more accurate simulation of various style characteristics and helps the model gradually learn and adapt to subtle differences between styles. Additionally, it exposes the model to a broader range of style features with different data distributions, thereby enhancing its generalization capability in unseen domains. The significant performance gains over five adverse-weather scenarios and the Real to Art benchmark demonstrate the superiorities of our method. 
\end{abstract}

%% file: sec/1_intro.tex
\vspace{-15pt}
\section{Introduction}
\vspace{-5pt}
\label{sec:intro}

\begin{figure}[t]
  \centering
  \includegraphics[width=0.9\linewidth]{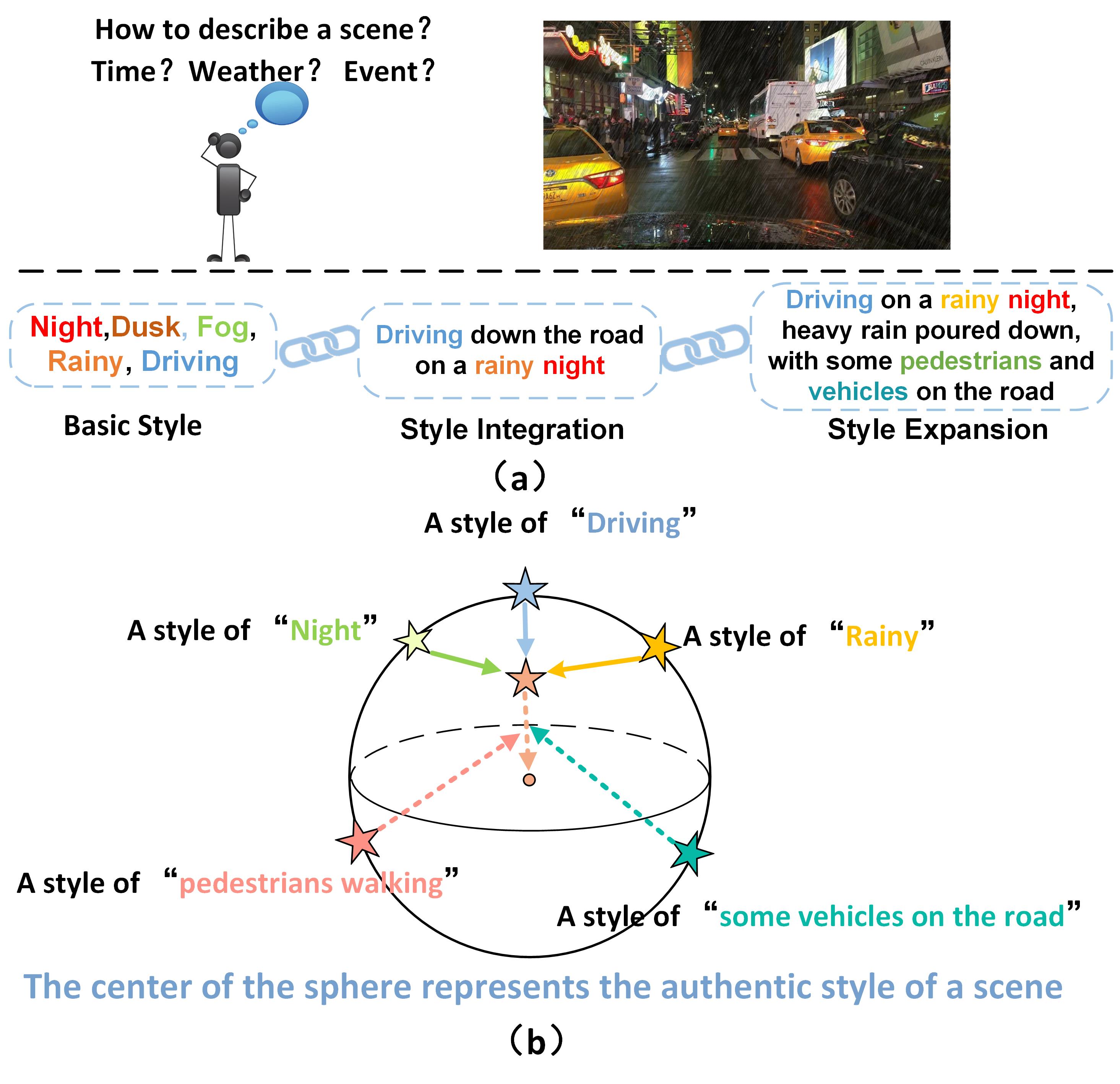}
  \vspace{-0.1in}
  \caption{Style evolution along Chain-of-Thought involves a process that progresses from coarse to fine, continuously refining and expanding styles. (a) illustrates the style description guided by the Chain-of-Thought. (b) demonstrates how using text prompts ranging from simple to complex guides style evolution, enabling the model to progressively learn and adapt to subtle differences between styles.}
    \label{F1}
    \vspace{-0.22in}
\end{figure}

Single-domain generalization (DG) \cite{domain1, domain2, domain3} aims to train a model on source domains to generalize to unseen target domains, posing a greater challenge than standard domain generalization. Though widely studied in image classification \cite{cc1, cc2, cc3}, it remains a developing area in object detection \cite{od3, od4}. Currently, there are two mainstream solutions. Firstly, decoupling domain-invariant features through self-supervised methods enhances the model's generalization ability in the target domain, such as \cite{S_DG, wu2023discriminating}. Secondly, some approaches utilize textual prompts to estimate cross-domain information, enhancing the model's generalization capability. Specifically, by employing a single textual prompt, these methods leverage the multimodal capabilities of vision-language models to simulate target domain styles \cite{Wpromptstyler, Poda} or estimate domain shifts \cite{C_Cap}, thereby improving the model's detection performance in the target domain. This approach typically employs a single-step textual prompt, hence referred to as the one-step prompt method. Presently, one-step prompt methods have achieved state-of-the-art performance in Single-domain generalization for image classification \cite{cc1, cc2, cc3}, object detection \cite{od3, od4, sg5}, and semantic segmentation \cite{s1, s2, sg4}.

Existing one-step prompt methods typically rely on single-step prompts to estimate cross-domain information. These prompts often contain basic, essential information, such as ‘driving on a rainy night’. For example, C-GAP \cite{C_Cap} and PODA \cite{Poda} utilize textual descriptions of weather and time to estimate domain shifts and simulate styles across different domains, aiming to improve the model’s generalization in unknown domains. However, in complex scenarios, such as combinations of styles like rainy and night, the performance of one-step prompts tends to be limited. This limitation arises because one-step prompts struggle to effectively synthesize combined information involving multiple styles. For instance, an image of driving on a rainy night includes multiple style features. Using one-step prompt methods for style generation in such cases often results in overly simplistic style effects, which fail to accurately capture the true style of real-world scenarios.

To address the limitation, we propose a new method, i.e., Style Evolving along Chain-of-Thought, which aims to progressively integrate and expand style information along the chain of thought, enabling continuous style evolution. This method not only provides a more accurate simulation of the true styles in real-world scenarios but also helps the model gradually learn and adapt to the subtle differences between various styles, thereby enhancing its generalization capability in unseen domains.
Specifically, as illustrated in Fig. \ref{F1} (a), we use text prompts that evolve from simple to complex, rough to refined, and partial to holistic to guide the style evolution. In the absence of target-domain data, this approach allows for the gradual accumulation and fusion of multiple styles while continuously expanding style diversity, as shown in Fig. \ref{F1} (b). Since the category space remains consistent between the target and source domains, with differences mainly in the style space, the model can progressively learn and adapt to the subtle differences between styles during the style evolution process. Additionally, exposing the model to a wide range of style features with different data distributions during training can robustly adapt to various styles, thereby improving its generalization ability in unknown domains. To more effectively simulate styles during training and minimize substantial semantic information loss, we decouple style features and exclusively perform style evolution of these style features. This approach effectively achieves style evolution while alleviating the impact of style transfer on class-related features.
Furthermore, we introduce a novel concept, the Class-Specific Prototype, designed to capture class-related semantic features. This prototype captures inherent invariant characteristics across different categories, thereby enriching semantic features. It also acts as a supervisory signal, guiding the disentanglement process more effectively and facilitating the separation of class-related features. The significant performance gains over five adverse-weather scenarios and the Real to Art benchmark demonstrate the superiorities of our method.

To summarize, our contributions are as follows:

• We propose a new method, i.e., Style Evolving along Chain-of-Thought, which aims to progressively simulate styles with different data distributions, enhancing the model's generalization capability in unseen domains.

• We achieve accurate style evolution by introducing feature disentanglement and Class-Specific Prototypes while avoiding the loss of class-related semantic information.

• We evaluate the efficacy of our proposed approach across five diverse weather-driving scenarios and Real to Art benchmarks, achieving state-of-the-art performance. 

%% file: sec/2_relatework.tex
\vspace{-5pt}
\section{Relate Work}

\label{sec:formatting}

\begin{figure*}[!t]
  \centering
  \includegraphics[width=2.0\columnwidth]{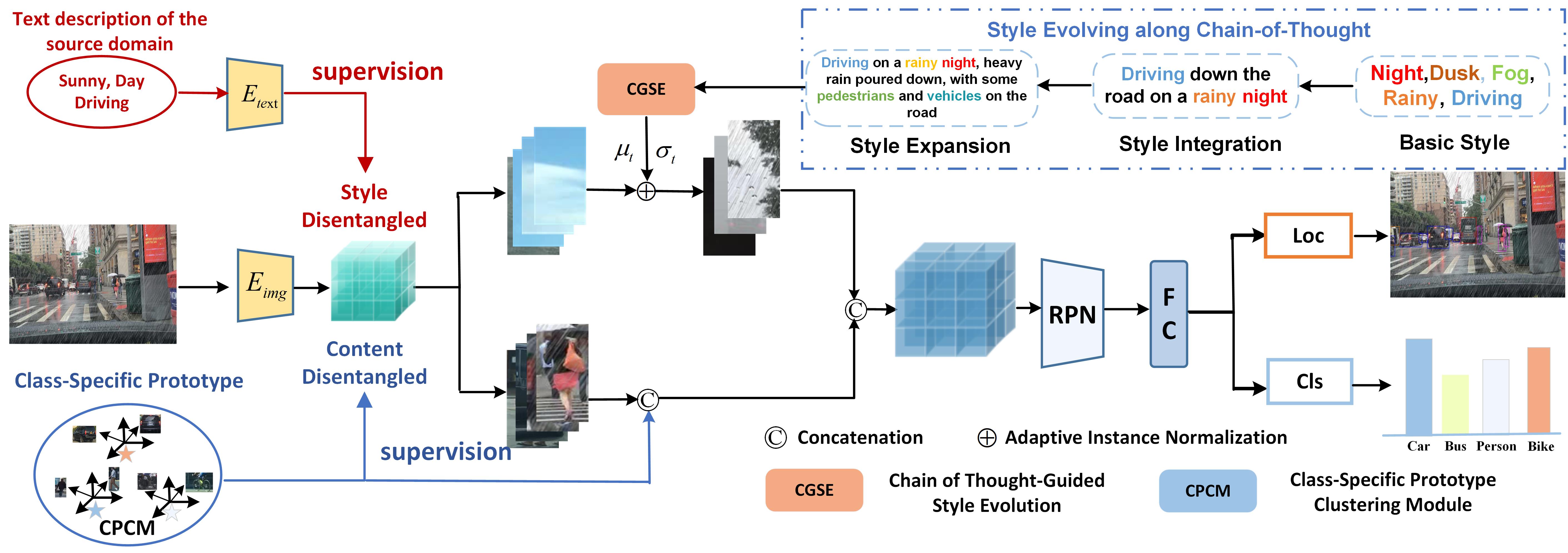}
  \caption{\textbf{The overall architecture of the model.} Leveraging contrastive loss and two consistency losses, we disentangle the first-layer features into content and style features. For style features, the CGSE module is employed to obtain style transfer parameters for style migration. Regarding content features, semantic enhancement is performed through class-specific prototypes. Ultimately, these two features are fused and input into the backbone network to generate features from layers 2 to 4. The output is then fed into the Region Proposal Network (RPN) for subsequent target localization and classification tasks.}
    \label{F2}
     \vspace{-0.1in}
\end{figure*}

\subsection{Single-Domain Generalization}
Single Domain Generalization (SDG), a more challenging research topic in the domain generalization field, can achieve strong generalization across multiple unseen target domains with training solely on a source domain. This makes SDG a precious research area, particularly for scenarios where obtaining data in extreme conditions is challenging. Currently, SDG research is well-established in image classification, such as \cite{C2, C3} which improves the model's generalization on the target domain by introducing data augmentation strategies and employing adversarial training to generate diverse input images. 
SDG is also explored in the field of semantic segmentation, such as \cite{Poda, sg2, sg3}. 
While SDG tasks have shown impressive results in image classification and semantic segmentation, they haven't been thoroughly investigated in the field of object detection. Single-DGOD \cite{S_DG} is the first method proposed for single-domain generalization tasks in object detection, contributing an excellent dataset and an effective method. It employs Cyclic-Disentangled Self-Distillation to extract domain-invariant and domain-specific features, enhancing generalization to the target domain. These methods enhance the generalization performance in the target domain through data augmentation or extracting domain-invariant features, without incorporating text prompts.
\vspace{-5pt}
\subsection{Prompt-driven Single-Domain Generalization}

The contrastive image-language pretraining model CLIP has recently achieved remarkable success in multimodal learning, supporting diverse downstream tasks like zero-shot image synthesis \cite{ZS1, ZS2}, multimodal fusion \cite{mf3}, semantic segmentation \cite{SS}, open-vocabulary object detection \cite{OVD}, and few-shot learning \cite{Fs}. Meanwhile, text prompt-based methods have achieved state-of-the-art performance in Single-domain generalization for image classification \cite{cc1, cc2, cc3}, object detection \cite{od3, od4}, and semantic segmentation \cite{s1, s2}. For instance, PODA \cite{Poda} achieves zero-shot domain adaptation in semantic segmentation through a prompting approach and exhibits promising results in the single-domain generalization domain. Most existing methods \cite{Poda, C_Cap} use simple text prompts, or one-step prompting, to obtain prior knowledge of the target domain. Although straightforward, this approach often fails to capture the target domain's complex characteristics and diverse styles. To address this issue, we propose a new method, i.e., Style Evolving along Chain-of-Thought, which aims to leverage the chain of thought to progressively simulate styles with different data distributions.

%% file: sec/3_method.tex
 \vspace{-5pt}
\section{Method}
 \vspace{-5pt}
This paper explores leveraging the multimodal capabilities of vision-language models to address single-domain generalization in object detection. The framework consists of two main components: the first, style evolution, is detailed in Section 3.1, while the second, style transfer, is illustrated in Fig. \ref{F2}.
\vspace{-5pt}
\subsection{Chain of Thought-Guided Style Evolution}
The parameters for style evolution are trained independently in a dedicated phase and repeated multiple times throughout the training process to ensure accurate style simulation. As shown in Fig. \ref{F3}, we use images from the source domain to generate image descriptions, achievable with any image captioning model. We enhanced a traditional image captioning model \cite{imagecaption} to generate only words pertinent to the image, selecting the most relevant terms such as ‘sunny,’ ‘day,’ and ‘realistic.’

We then leverage ChatGPT \cite{gpt} to generate supplementary descriptions from these keywords, creating distinct vocabularies for weather descriptions like ‘rainy’ and ‘foggy,’ and style descriptions like ‘anime’ and ‘art.’ This set is denoted as \( W = \{W_1, W_2, \ldots, W_n\} \), where \( n \) represents the number of vocabularies. Specifically, we construct five vocabularies: weather, time, style, action, and detail. From each vocabulary, we randomly select one word, denoted as \( W_r^i \), and then use CLIP for text feature extraction and initial style fusion. This process yields the foundational textual features for the initial stage of style evolution, \( F_t^1 \), as illustrated below:
\begin{eqnarray}
F_t^1 = \sum_{i=1}^{n} E_{text}(W_r^i).
\end{eqnarray}
Subsequently, we use ChatGPT\cite{gpt} to combine these words into phrases, denoted as \( P_r \). For example, ‘Driving down the road on a rainy night’ and ‘An art-style portrait image.’ We also integrate multiple styles to create a complex description that seamlessly fits into real-world scenarios. Next, we extract features from the composed description \( P_r \) and add them to the text features \( F_t^1 \) obtained in the first step:
\begin{equation}F_t^2=E_{text}(P_r)+F_t^1.\end{equation}
Next, we incorporate detailed stylistic descriptions into the existing phrases, obtaining a complete sentence $S_r$ describing the scene. This allows for style expansion based on the existing styles, simulating a richer variety of style features with different data distributions. Following this, feature extraction is conducted on $S_r$, and the outcomes are amalgamated with the previously extracted features $F_t^2$. This iterative process yields increasingly refined textual features $F_t^3$ for guiding subsequent style evolution:
\begin{figure}[t]
  \centering
  \includegraphics[width=0.9\linewidth]{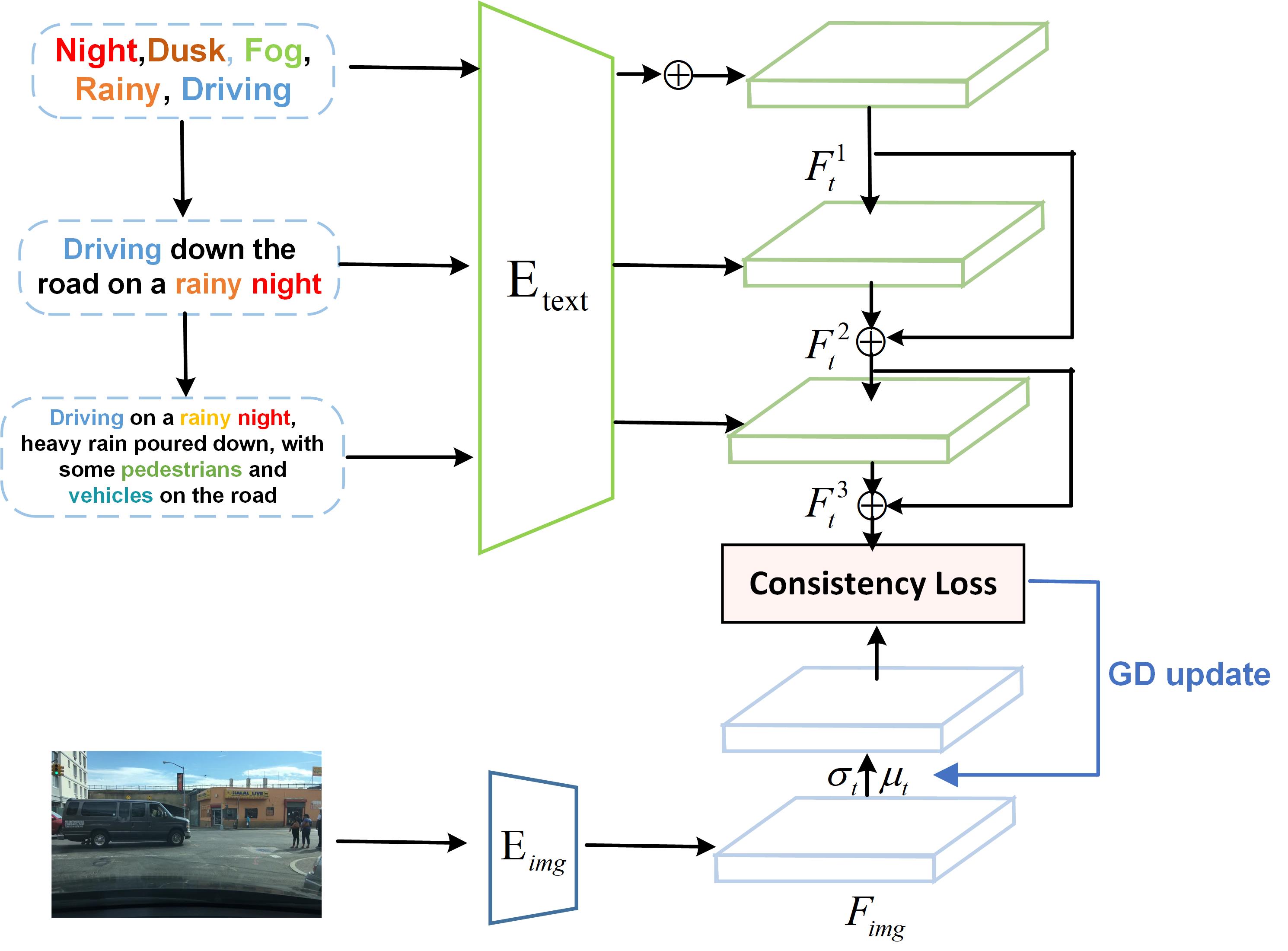}
  \caption{\textbf{Complex Style Evolution:} By using text prompts that progress from simplicity to complexity, the style is continuously evolved and expanded, thereby simulating a richer variety of style features with different data distributions. Parameters $\mu_t$ and $\sigma_t$ are trained using consistency loss, aligning the text features with the source domain visual features, and are subsequently utilized for style evolution.}
  \vspace{-0.2in}
    \label{F3}
\end{figure}
\begin{equation}F_t^3=E_{text}(S_r)+F_t^2.\end{equation}
After obtaining refined textual features $F_t^3$, we employ a visual feature extractor to extract source features $F_s$ from source domain images. By utilizing the first-layer features of the backbone, we perform style evolution. Here, we adopt the concept of Adaptive Instance Normalization (AdaIN) \cite{ada} for style evolution. Initially, source features $F_s$ undergo normalization, resulting in normalized features $F_s^\prime$, where $\mu(F_s)$ and $\sigma(F_s)$ represent the mean and variance calculations along the channel dimension of the source domain feature $F_s$:
\begin{equation} F_s^\prime=\frac{F_s-\mu(F_s)}{\sigma(F_s)}.\end{equation}

Subsequently, we utilize textual features to train two learnable parameters $\mu_t$ and $\sigma_t$ for performing style evolution on the visual features $F_s^\prime$ of the source domain. Performing style evolution using source domain visual features allows the model to encounter style features with different data distributions during the training phase, enhancing the feature's generalization across different domains:
\begin{equation}F_i=\sigma_t(F_s^{\prime})+\mu_t.\end{equation}

By calculating the cosine similarity along the channel dimension between textual features $F_t^3$ and visual features $F_i$, we compute the consistency loss $\mathcal{L}_\mathrm{tc}$:
\begin{equation}\mathcal{L}_{tc}=1-sim(F_i,F_t^3),\end{equation}
 where $\mathrm{sim(a,b)}$ denotes the average of the cosine similarity between all corresponding elements of feature maps a and b. 
Through iterative updates, the two parameters $\mu_t$ and $\sigma_t$, are continuously refined. In the end, we obtain the parameters for source domain image style evolution. 

\subsection{Style Disentangled Module}
To avoid losing significant semantic information during the style transfer process, we initially decouple the features of the first layer and then perform style evolution exclusively on the style features. We have designed two feature extractors $E_{Style}$ and $E_{Content}$ for extracting style features $F_s$ and content features $F_c$ separately. Since style evolution is conducted on the first-layer features, here $F_1$ represents the first-layer feature map. The process is as follows:
\begin{equation}F_s=E_{Style}\left(F_1\right),\ {\ F}_c=E_{Content}\left(F_1\right).\end{equation}

To better ensure the extraction of style and content features and to ensure their more thorough decoupling, we employ three types of losses for supervision. Firstly, we use contrastive loss $\mathcal{L}_\mathrm{d}$ to ensure a more thorough decoupling of style and content features. For the sake of simplicity in notation, let $\mathrm{sim(a,b)}$ denote the average of the cosine similarity between all corresponding elements of feature maps a and b. The contrastive loss $\mathcal{L}_\mathrm{d}$ is defined as follows:
\begin{equation}\mathcal{L}_\mathrm{d}=-(\log\frac{\exp(sim(F_1,F_s)/\mathfrak{\tau})}{\sum_{j=0}^1\exp\left(sim(F_1,P[j])/\mathfrak{\tau}\right)}),\end{equation}
where $P=[F_s, F_c]$, and $\mathfrak{\tau}$ is a hyperparameter. In the experiments,  $\mathfrak{\tau}$ is set to 1.0. Additionally, we utilize textual prompts $F_t^s$ from the source domain to supervise decoupling the style features, aiming to decouple the style features $F_s$ consistent with the source domain text descriptions $F_t^s$. We apply consistency loss $\mathcal{L}_\mathrm{sc}$ to constrain the feature decoupling, using it to represent the textual features of the source domain descriptions and to denote the decoupled style features $F_s$. The consistency loss is computed between these two representations $F_s$ and $F_t^s$. The consistency loss $\mathcal{L}_\mathrm{sc}$ is defined as follows:
\begin{equation}\mathcal{L}_{sc}=1-sim\left(F_s,F_t^s\right).\end{equation}

Regarding content features, we aim to incorporate more class-specific information. Therefore, we introduce a novel prototype, the class-specific prototype, to cluster class-specific relevant features. Detailed descriptions of this approach will be provided in Section \ref{sec3.3}. In this section, we obtain the Class-Specific Prototype \( F_p \) through clustering high-level features and then downsample it to the same dimension as \( F_c \) to supervise the decoupling of content features, as shown below:
\begin{equation}\mathcal{L}_{gc}=1-sim\left(Down(F_p),F_c\right),\end{equation}
where \( Down \) represents the downsampling operation. This ensures that the content features include more class-related semantic characteristics.

\subsection{Class-Specific Prototype Clustering Module}
\label{sec3.3} 

The acquisition of Class-Specific prototypes is optimized by capturing distinctive invariant features for each class. We employ the Class-Specific Prototype Clustering Module to effectively cluster Class-Specific prototypes, denoted as \( F_p \). These prototypes act as supervisory signals, aiding in style decoupling and enhancing content features.

The Class-Specific Prototype Clustering Module begins with 2-norm normalization of the input feature map \( F \), followed by dimensionality reduction through convolution. The convolutional layer calculates scores, known as soft assignment probabilities, for each pixel in the feature map relative to a set of \( K \) prototypes. These probabilities are then normalized using the SoftMax function, resulting in \(\theta\):
\begin{eqnarray}
\theta=ConvSoft(L2(F)).
\end{eqnarray}

The number of Class-Specific Prototypes denoted as \( K \), is a hyperparameter matching the number of classes in the dataset. Dimensionality reduction aligns the number of channels with \( K \). Scores for each pixel, corresponding to \( K \) prototypes, are converted into a tensor with the same spatial dimensions as the input feature map. The input feature map is reshaped to (N, C, H * W). During initialization, a learnable 2D tensor of cluster centers with dimensions (K, C) is created and expanded to (N, K, C), referred to as \( sp \). Residuals between each prototype and pixel are computed and weighted by the soft assignment probabilities. The calculation process is as follows:
\begin{eqnarray}
F_p^1=\theta(Resize\left(\theta\right)-Resize\left(sp\right)).
\end{eqnarray}

The residuals are summed to produce the weighted residuals for each prototype, resulting in a tensor with shape (N, K, C). After applying 2-norm normalization, the weighted residuals are reshaped to (N, K * C) along the final dimension. The process is as follows:
\begin{eqnarray}
F_p^2=L2(Resize(L2\left(sum\left(F_p^1\right)\right))).
\end{eqnarray}

The weighted residuals $F_p^2$  are then passed to the fully connected layer, and the output tensor is reshaped to (N, K, C) and then expanded to the original spatial dimensions of the input feature map to obtain  $F_p^3$, which is then returned along with the cluster center centroid:
\begin{eqnarray}
F_p^3=Resize(Linear\left(F_p^2\right)).
\end{eqnarray}

Furthermore, the $F_p^3$ is concatenated with the input feature map $F$ along the channel dimension and then reduced back to its original channel dimension using convolution: 
\begin{eqnarray}
F_p=Conv(Cat\left(F,F_p^3\right)),
\end{eqnarray}
this results in the prototype-enhanced feature map $F_p$, which better highlights the features of the foreground.

%% file: sec/4_experiment.tex
\vspace{-2pt}
\section{Experiments}
In the experiments, for SDG, we follow the settings of the work \cite{S_DG} to validate the model's generalization capability. Additionally, to further verify the effectiveness of our method, we evaluated the model on the Reality-to-Art generalization benchmark.
\vspace{-5pt}
\subsection{Experimental setup}
\textbf{Dataset. }
The entire experiment is structured around two benchmarks. \textbf{Diverse
Driving Weather Scenarios.} For training, we exclusively use a dataset \cite{S_DG} of daytime sunny images, comprising 19,395 images and 8,313 images for validation and subsequent model selection. After training the model on the source domain data, we directly test its performance on the target domain, which includes four distinct weather conditions: 26,158 images of clear night scenes, 3,501 images of rainy scenes at dusk, 2,494 images of rainy scenes at night, and 3,775 images of foggy scenes during the daytime. The category space is consistent across all five datasets, including seven classes: bus, bicycle, car, motor, person, rider, and truck. 

\textbf{Generalization from Reality to Art.} During training, as in \cite{Art}, the Pascal VOC2007 and VOC2012 trainval sets are used as the training datasets. The method's generalization ability to unseen domains is evaluated on the Clipart1k, Watercolor2k, and Comic2k datasets. The Clipart dataset shares the same 20 classes as Pascal VOC, while Watercolor2k and Comic2k consist of 6 classes each, which are subsets of the Pascal VOC classes.

\textbf{Metric.}
For a fair comparison with other methods, we employ the same evaluation metric parameters as other S-DGOD methods, measuring our model's performance using the mean Average Precision (mAP) metric with an Intersection-over-Union (IoU) threshold of 0.5. For different test datasets, we assess the mAP@0.5 for all categories in the dataset to evaluate the generalization performance of our model across various scenes and classes.
\vspace{-5pt}
\subsection{Implementation Details}

To evaluate the effectiveness of our method, we use the same baseline model, Faster R-CNN \cite{ren2015faster}, as employed in other S-DGOD approaches. To thoroughly assess the model's generalization capability, we present detection results using three different backbones, as shown in Table \ref{tab1} and Table \ref{t2}. The detector is initialized with pre-trained weights from CLIP \cite{CLIP}, and during training, the weights of layers 1 through 3 remain frozen.

 \begin{table}[!t]
  \centering
\vspace{-0.05in}
  \caption{Single-Domain Generalization Results (mAP(\%)) in Diverse Weather Driving Scenarios. \textbf{The bold sections} represent the best results.}
    \vspace{-0.1in}
  \resizebox{\linewidth}{!}{
    \begin{tabular}{cc|c|cccc}
    \toprule
    \toprule
    \multicolumn{2}{c|}{\multirow{2}[4]{*}{Method}} & \multicolumn{1}{c}{} &       & \multicolumn{1}{c}{mAP} &       &  \\
\cmidrule{3-7}    \multicolumn{2}{c|}{} & \multicolumn{1}{c|}{Day Clear} & \multicolumn{1}{c}{Night Sunny} & \multicolumn{1}{c}{Dusk Rainy} & \multicolumn{1}{c}{Night Rainy} & \multicolumn{1}{c}{Day Foggy} \\
    \midrule
    \multicolumn{2}{c|}{Faster R-CNN(Res101) \cite{ren2015faster}} & 48.1  & 34.4  & 26.0  & 12.4  & 32.0  \\
    \multicolumn{2}{c|}{SW \cite{SW}} & 50.6  & 33.4  & 26.3  & 13.7  & 30.8  \\
    \multicolumn{2}{c|}{IBN-Net \cite{IBN}} & 49.7  & 32.1  & 26.1  & 14.3  & 29.6  \\
    \multicolumn{2}{c|}{IterNorm \cite{Interfrom}} & 43.9  & 29.6  & 22.8  & 12.6  & 28.4  \\
    \multicolumn{2}{c|}{ISW \cite{ISW}} & 51.3  & 33.2  & 25.9  & 14.1  & 31.8  \\
    \multicolumn{2}{c|}{S-DGOD (Res101)\cite{S_DG}} & 56.1 & 36.6  & 28.2  & 16.6  & 33.5  \\
    \multicolumn{2}{c|}{C-Gap (Res101)\cite{C_Cap}} & 51.3  & 36.9  & 32.3  & 18.7  & 38.5  \\
    \multicolumn{2}{c|}{PDOC (Res101)\cite{pdoc}} & 53.6  & 38.5  & 33.7  & 19.2  & 39.1  \\
    \multicolumn{2}{c|}{UFR (Res101)\cite{UFR}} & 58.6  & 40.8  & 33.2  & 19.2  & 39.6  \\
    \multicolumn{2}{c|}{DIV (Res101) \cite{Div}} & 52.8  & 42.5  & 38.1  & 24.1  & 37.2  \\
    \midrule
    \multicolumn{2}{c|}{Ours(Res50)} & 48.4 & 33.5  & 28.6  & 16.1  & 36.9  \\
   
     \multicolumn{2}{c|}{Ours(Res101)} &  55.4  & 42.0  & 39.2  & 24.5  & 40.6\\
    \multicolumn{2}{c|}{Ours(Swin)} & \textbf{64.4} & \textbf{52.7} & \textbf{49.5} & \textbf{33.7} & \textbf{44.9} \\
    \bottomrule
    \bottomrule
    \end{tabular}%
    
    }
    \vspace{-15pt}
  \label{tab1}%
\end{table}%

\textbf{Style Evolution Stage.} We enhanced a traditional image captioning model \cite{imagecaption} to generate only words related to the image and select the most relevant terms, such as ‘sunny,’ ‘day,’ and ‘realistic.’ Any image captioning model can be used as a substitute here. Then we use ChatGPT \cite{gpt} to generate related words, forming multiple vocabularies. We construct five vocabularies: weather, time, style, action, and detail. In the first phase of style evolution, we randomly select a word from multiple vocabularies, such as ‘rainy’ or ‘anime,’ and use a text encoder to extract features, obtaining initial text features. In the second phase, we leverage ChatGPT to generate phrases, such as ‘driving down the road on a rainy night’ or ‘an anime-style image.’ Finally, we extract words from the detail vocabulary, such as ‘pedestrian’ and ‘motorcycle,’ to further enrich the descriptions, creating more refined style texts. These style text features are combined with the features obtained in the second phase to achieve progressively evolved style features. Using the final style features, we learn two sets of style parameters, denoted as \(\mu_t\) and \(\sigma_t\), each consisting of 256 learnable channels. We use the stochastic gradient descent optimizer for parameter optimization, with a learning rate of 1.0, a momentum of 0.9, and a weight decay of 0.0005. Training is conducted on a single 3090 GPU with a batch size of 2.

\textbf{Style Transfer Stage.} In each training iteration, we randomly select one set of learned style parameters from multiple sets for style transfer, simulating the style space of the target domain. For the decoupled style features, we apply style transfer using randomly selected style parameters $\mu_t$ and $\sigma_t$. Additionally, we utilize Class-Specific prototypes for semantic enhancement of the content features. After obtaining the transferred style features and enhanced content features, we merge them and input them into the backbone network for feature extraction at layers 2-4. Subsequently, we pass them through the Region Proposal Network (RPN) for subsequent classification and localization tasks. Ablation study analysis on other feature layers and specific examples of stylistic prompts can be found in the supplementary materials.

\begin{figure*}[t]
  \centering
  \vspace{-0.2in}
  \includegraphics[width=2.0\columnwidth]{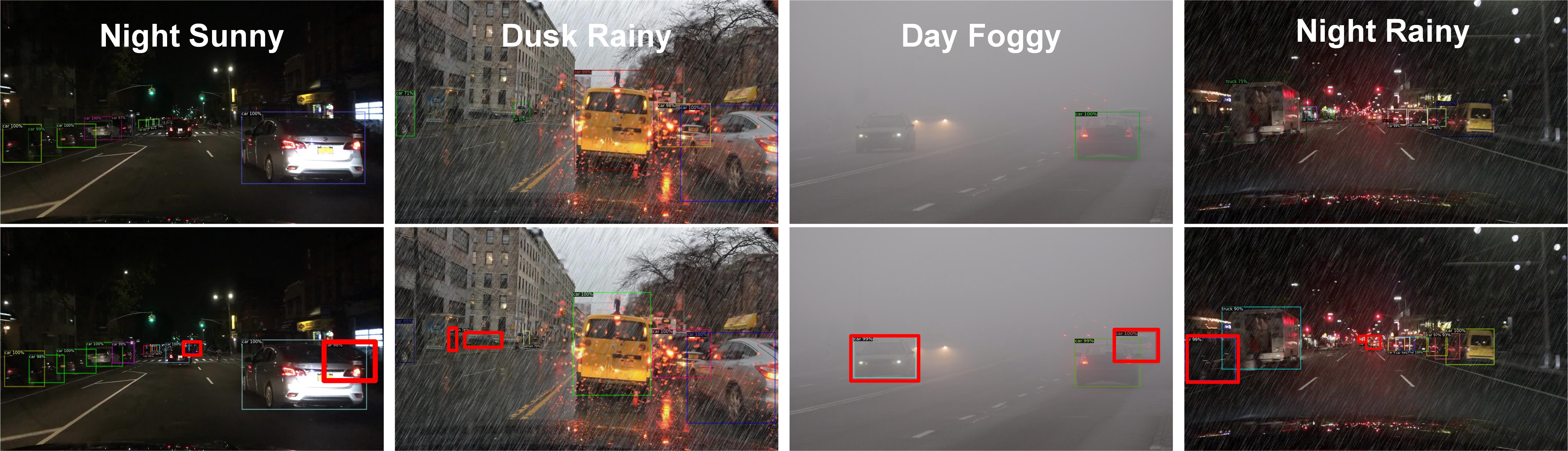}
  \caption{\textbf{Qualitative Results}: Detection results on different weather conditions. The first and second rows display the results from C-GAP \cite{C_Cap} and our method. Objects highlighted in \textbf{red boxes} represent those missed by C-GAP \cite{C_Cap} but correctly identified by our method.}
    \label{F5}
    \vspace{-0.15in}
\end{figure*}


\vspace{-0.1in}
\subsection{Comparison with the State of the Art}

\begin{figure*}[t]
  \centering
  \includegraphics[width=2.0\columnwidth]{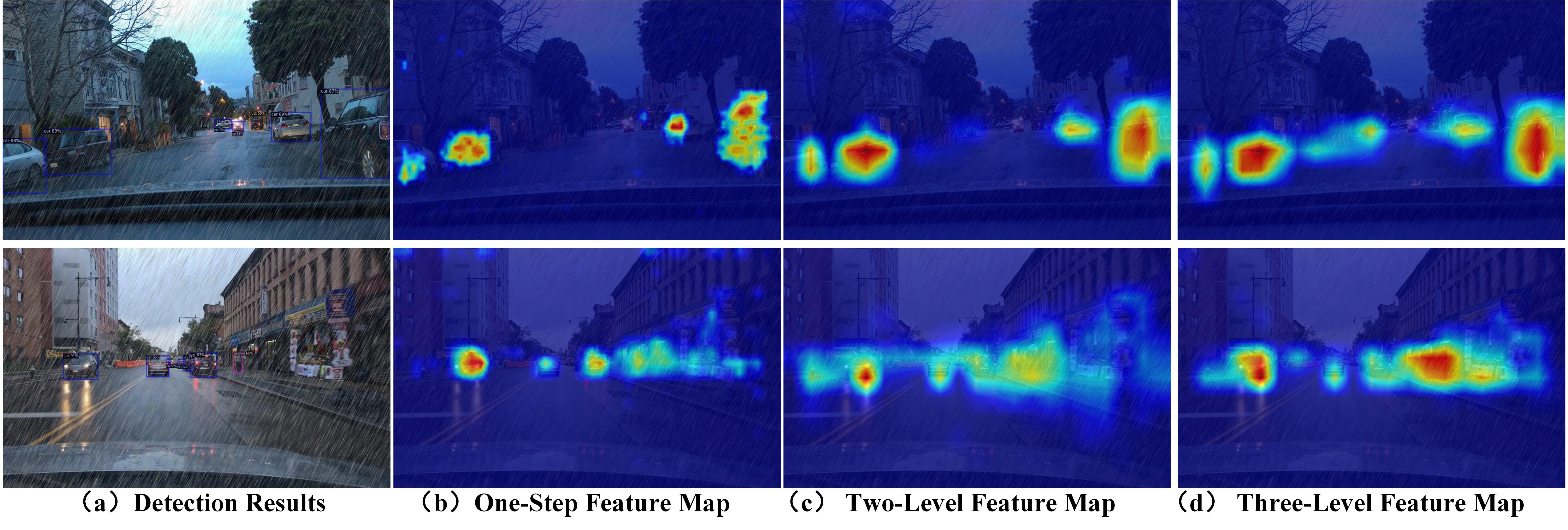}
  \vspace{-0.1in}
  \caption{\textbf{Visualization analysis of our method}: The first column shows the model's detection results. The second column visualizes the features with style evolution using a one-step method, while the third and fourth columns represent style evolution guided by two-level and three-level chains of thought, respectively. The feature heatmaps show that as the chains of thought progress, the model increasingly focuses on foreground objects, validating the effectiveness of the method. }
    \label{vis}
     \vspace{-0.15in}
\end{figure*}

\textbf{Diverse Driving Weather Scenarios.} We compared our method with the current state-of-the-art S-DGOD method, as shown in Table \ref{tab1}. Using the same backbone network, our method achieved the best results across four unknown target domains. Additionally, we reported the model's detection performance with different backbone networks. When using SwinTransformer as the backbone, the accuracy of our method was further improved. Figure \ref{vis} shows the detection results in target domains. The results indicate that our model excels under challenging and low-visibility conditions, such as rainy nights, accurately identifying objects that are difficult for the human eye to discern.

\textbf{Generalization from Reality to Art.} We evaluate our method on the challenging domain shift from reality to artistic benchmark, where the distribution shift is substantial, particularly from PASCAL VOC to Clipart, Watercolor, and Comic. As shown in Table \ref{t2}, our model, using the same ResNet-101 backbone, achieves the best performance across these three target domains, thereby validating its generalization capability to unknown target domains. Compared to the one-step prompt method C-Gap, our method improves performance by 18\%, 13\%, and 9\% on the Comic, Watercolor, and Clipart datasets, respectively. This demonstrates that our approach facilitates better style evolution, enabling the model to handle a broader range of data distributions during training and thereby enhancing its generalization performance on unseen domains.

\begin{table}[t]
  \centering
   \vspace{-0.05in}
  \caption{Single-Domain Generalization Results (mAP(\%)) from Real to Artistic. \textbf{The bold} represent the best results.}
  \resizebox{\linewidth}{!}{
    \begin{tabular}{cc|c|ccc}
    \toprule
    \toprule
    \multicolumn{2}{c|}{\multirow{2}[4]{*}{Method}} & \multicolumn{1}{c}{} &       & \multicolumn{1}{c}{mAP} &  \\
\cmidrule{3-6}    \multicolumn{2}{c|}{} & \multicolumn{1}{c|}{VOC} & \multicolumn{1}{c}{Comic} & \multicolumn{1}{c}{Watercolor} & \multicolumn{1}{c}{Clipart} \\
    \midrule
    \multicolumn{2}{c|}{Faster R-CNN(Res101)\cite{ren2015faster}} & 80.4  & 19.4  & 45.6  & 26.5  \\
    \multicolumn{2}{c|}{NP(Res101)\cite{np}} & 79.2  & 28.9  & 53.3  & 35.4  \\
    \multicolumn{2}{c|}{C-Gap(Res101)\cite{C_Cap}} & 80.5  & 29.4  & 50.7  & 36.7  \\
    \multicolumn{2}{c|}{DIV\cite{Div}} & 80.1  & 33.2  & 57.4  & 38.9  \\
    \midrule
    \multicolumn{2}{c|}{Ours(Res50)} & 78.5  & 29.4  & 53.2  & 34.7  \\
    \multicolumn{2}{c|}{Ours(Res101)} & 82.9  & 34.8  & 57.5  & 40.2  \\
    \multicolumn{2}{c|}{Ours(Swin)} & \textbf{87.6} & \textbf{36.9} & \textbf{60.7} &\textbf{42.5}\\
    \bottomrule
    \bottomrule
    \end{tabular}%
    }
  \label{t2}%
  \vspace{-15pt}
\end{table}%

\begin{table}[!h]
  \centering
  \caption{Per-class results (\%) on Day Foggy. \textbf{The bold sections} represent the best results.}
  \vspace{-0.1in}
  \fontsize{8}{6}\selectfont
  \resizebox{\linewidth}{!}{
    \begin{tabular}{cc|ccccccc|c}
    \toprule
    \toprule
     \multicolumn{2}{c|}{\multirow{2}[4]{*}{Method}} & \multicolumn{7}{c|}{AP}                     & \multicolumn{1}{c}{mAP} \\
\cmidrule{3-10}    \multicolumn{2}{c|}{} & \multicolumn{1}{c|}{Bus} & \multicolumn{1}{c|}{Bike} & \multicolumn{1}{c|}{Car} & \multicolumn{1}{c|}{Motor} & \multicolumn{1}{c|}{Person} & \multicolumn{1}{c|}{Rider} & \multicolumn{1}{c|}{Truck} & \multicolumn{1}{c}{All} \\
    \midrule
    \multicolumn{2}{c|}{Faster R-CNN \cite{ren2015faster}} & 28.10  & 29.70  & 49.70  & 26.30  & 33.20  & 35.50  & 21.50  & 32.00  \\
    \multicolumn{2}{c|}{S-DGOD \cite{S_DG}} & 32.90  & 28.00  & 48.80  & 29.80  & 32.50  & 38.20  & 24.10  & 33.50  \\
    \multicolumn{2}{c|}{C-Gap \cite{C_Cap}} & 36.10  & 34.30  & 58.00  & 33.10  & 39.00  & 43.90  & 25.10  & 38.50  \\
     \multicolumn{2}{c|}{PDOC \cite{pdoc}} & 36.10  & 34.50  & 58.40  & 33.30  & 40.50  & 44.20  & 26.20  & 39.10  \\
      \multicolumn{2}{c|}{UFR \cite{UFR}} & 36.90  & 35.80  & 61.70 & 33.70 & 39.50  & 42.20  & 27.50  & 39.60  \\
      
    \midrule
    \multicolumn{2}{c|}{Ours(Res101)} & 39.33 & 36.28 & 60.82 & 33.79 & 39.41 & 42.68  & 31.77 & 40.58 \\
      \multicolumn{2}{c|}{Ours(Swin)} & \textbf{42.11} & \textbf{38.32} & \textbf{62.79} & \textbf{40.75} &\textbf{47.48} & \textbf{48.04} & \textbf{35.41} & \textbf{44.99} \\
    \bottomrule
    \bottomrule
    \end{tabular}%
    }
    \vspace{-0.1in}
  \label{table2}%
\end{table}%

\textbf{Daytime Clear to Day Foggy.}  Compared to clear weather, foggy conditions introduce even more severe occlusion, especially in dense fog scenarios. The performance of different object categories varies significantly under dense fog, posing a significant challenge to detection tasks. As shown in Table \ref{table2}, our model outperforms all existing state-of-the-art single-domain generalization object detection methods, including C-Gap \cite{C_Cap}, PDOC \cite{pdoc}, and UFR \cite{UFR} across almost all categories. This remarkable achievement not only further demonstrates our superior generalization performance in foggy conditions but also validates the robustness of our model in extreme scenarios. Since the DIV\cite{Div} method does not report the AP values for the seven categories under different weather conditions, we are unable to make a comparison with it.

\textbf{Daytime Clear to Dusk Rainy.} 
From Fig. \ref{vis}, it can be observed that in dusk rainy conditions, raindrops cause partial occlusion and the insufficient lighting further increases the overall detection difficulty. As shown in Table \ref{tab3}, our method significantly outperforms other approaches in the Dusk Rainy scenario, demonstrating its superiority in extreme conditions. Per-class results for the other two weather conditions are available in the supplementary materials.

\begin{table}[!t]
  \centering
  \vspace{-5pt}
  \caption{Per-class results  (\%) on Dusk Rainy. \textbf{The bold sections} represent the best results.}
  \fontsize{8}{6}\selectfont
  \resizebox{\linewidth}{!}{
    \begin{tabular}{cc|ccccccc|c}
    \toprule
    \toprule
    \multicolumn{2}{c|}{\multirow{2}[4]{*}{Method}} & \multicolumn{7}{c|}{AP}                     & \multicolumn{1}{c}{mAP} \\
\cmidrule{3-10}    \multicolumn{2}{c|}{} & \multicolumn{1}{c|}{Bus} & \multicolumn{1}{c|}{Bike} & \multicolumn{1}{c|}{Car} & \multicolumn{1}{c|}{Motor} & \multicolumn{1}{c|}{Person} & \multicolumn{1}{c|}{Rider} & \multicolumn{1}{c|}{Truck} & \multicolumn{1}{c}{All} \\
    \midrule
    \multicolumn{2}{c|}{Faster R-CNN \cite{ren2015faster}} & 28.50  & 20.30  & 58.20  & 6.50  & 23.40  & 11.30  & 33.90  & 26.00  \\
    \multicolumn{2}{c|}{S-DGOD \cite{S_DG}} & 37.10  & 19.60  & 50.90  & 13.40  & 19.70  & 16.30  & 40.70  & 28.20  \\
    \multicolumn{2}{c|}{C-Gap \cite{C_Cap}} & 37.80  & 22.80  & 60.70  & 16.80  & 26.80  & 18.70  & 42.40  & 32.30  \\
    \multicolumn{2}{c|}{PDOC \cite{pdoc}} & 39.40  & 25.20  & 60.90  & 20.40  & 29.90  & 16.50  & 43.90  & 33.70  \\
    \multicolumn{2}{c|}{UFR \cite{UFR}} & 37.10  & 21.80  & 67.90 & 16.40  & 27.40  & 17.90  & 43.90  & 33.20  \\
    \midrule
    \multicolumn{2}{c|}{Ours(Res101)} & 45.75 & 19.20 & 74.32 & 23.43& 35.81 & 22.14 & 53.47 & 39.16 \\
    \multicolumn{2}{c|}{Ours(Swin)} & \textbf{57.34} & \textbf{40.23} & \textbf{76.76} &\textbf{28.36} &\textbf{50.03} & \textbf{32.66} & \textbf{60.80} & \textbf{49.45} \\
    \bottomrule
    \bottomrule
    \end{tabular}%
    }
  \label{tab3}%
  \vspace{-15pt}
\end{table}%

\textbf{Generalization from Reality to Comic.} We evaluate the generalization ability of our method in transferring from real domains to artistic domains, where the data distribution differences are more significant compared to various weather datasets. This includes generalization scenarios from PASCAL VOC to Clipart, Watercolor, and Comic datasets.
Table \ref{comic} shows the generalization results from PASCAL VOC to the Comic style dataset. As observed from the table, using the same Res101 backbone network, our method achieves an 18\% improvement on the Comic style dataset. This validates the generalization capability of our approach across different style target domains. Our study demonstrates that employing a Chain-of-Thought approach to guide style evolution enables the model to gradually learn and adapt to subtle differences between styles during the evolution process. Additionally, by simulating a wide range of style features with varying data distributions during training, we significantly enhance the model's robustness and generalization ability in unknown domains.

\begin{table}[htbp]
  \centering
  \vspace{-5pt}
  \caption{Per-class results (\%) on VOC to Comic. \textbf{The bold sections} represent the best results.}
   \fontsize{8}{6}\selectfont
  \resizebox{\linewidth}{!}{
    \begin{tabular}{cc|cccccc|c}
    \toprule
    \toprule
    \multicolumn{2}{c|}{\multirow{2}[4]{*}{Method}} & \multicolumn{6}{c|}{AP}               & \multicolumn{1}{c}{mAP} \\
\cmidrule{3-9}    \multicolumn{2}{c|}{} & \multicolumn{1}{c|}{Bike} & \multicolumn{1}{c|}{Bird} & \multicolumn{1}{c|}{Car} & \multicolumn{1}{c|}{Cat} & \multicolumn{1}{c|}{Dog} & \multicolumn{1}{c|}{Person} & \multicolumn{1}{c}{All} \\
    \midrule
    \multicolumn{2}{c|}{Faster R-CNN\cite{ren2015faster}} & 36.50  & 8.60  & 25.90  & 9.20  & 10.80  & 25.20  & 19.37  \\
    \multicolumn{2}{c|}{NP\cite{np}} & 42.44   & 18.25  & 38.79  & 17.33  & 24.29  & 32.18  & 28.88  \\
    \multicolumn{2}{c|}{C-Gap\cite{C_Cap}} & 41.98  & 15.94  & 41.97  & 18.77  & 24.60  & 33.26 & 29.42  \\
     \multicolumn{2}{c|}{DIV\cite{Div}} & \textbf{54.10} & 16.90 & 30.10 & \textbf{25.00} & 27.40 & 45.90 & 33.20 \\
    \midrule
    \multicolumn{2}{c|}{Ours(Res101)} & 47.00  & \textbf{22.99}  & 44.95  & 20.91  & 29.42  & 43.63  & 34.82  \\
    \multicolumn{2}{c|}{Ours(Swin)} & 47.73 & 21.72 & \textbf{47.60} & 23.27 & \textbf{30.78} & \textbf{50.42} & \textbf{36.92} \\
    \bottomrule
    \bottomrule
    \end{tabular}%
    }
  \label{comic}%
  \vspace{-15pt}
\end{table}%

\subsection{Ablation Study}

To validate the effectiveness of each module in our method, we performed ablation studies. The framework consists of several components: Chain-of-Thought-Guided Style Evolution, Style Disentanglement Module, and Class-Specific Prototype Clustering Module. We also compared our Chain-of-Thought approach with a one-step generation method to highlight its advantages.
Table \ref{teb6} shows the baseline method in the first row. The second row illustrates the one-step prompt method, where a complete prompt is provided, such as ‘Driving on a rainy night, heavy rain pouring down, with some pedestrians and vehicles on the road.’ The third row demonstrates Chain-of-Thought-Guided Style Evolution. The fourth row features our Style Disentanglement Module, which enhances style features. The fifth row incorporates Class-Specific Prototypes, further improving content features and supervising the disentanglement process.
The results confirm that the Style Disentanglement Module and Class-Specific Prototypes are effective and that Chain of Thought significantly improves generalization to unseen domains compared to the one-step prompt approach.

\begin{table}[htbp]
  \centering
  
  \caption{Ablation analysis (\%) of our proposed method. One-step represents the method using single-step prompts for style generation. CGSE represents the approach employing a chain of thought-guided styles evolving. SDM denotes style disentanglement, while CPCM signifies class-specific prototype semantic enhancement.}
 
    \resizebox{\linewidth}{!}{
    \begin{tabular}{ccccc|c|cccc}
    \toprule
    \toprule
    \multicolumn{5}{c|}{Method} & \multicolumn{1}{c|}{Source} & \multicolumn{4}{c}{Target} \\
    \midrule
    \multicolumn{1}{c}{Baseline} & \multicolumn{1}{c}{One-step} & \multicolumn{1}{c}{CGSE} & \multicolumn{1}{c}{SDM} & \multicolumn{1}{c|}{CPCM} & \multicolumn{1}{c|}{Day Clear} & \multicolumn{1}{c}{Night Clear} & \multicolumn{1}{c}{Dusk Rainy} & \multicolumn{1}{c}{Night Rainy} & \multicolumn{1}{c}{Day Foggy} \\
    \midrule
    $\checkmark$      &       &       &       &       & 49.6  & 34.7  & 25.7  & 11.8  & 28.4 \\
    $\checkmark$      & $\checkmark$      &       &       &       & 52.4  & 36.9  & 28.9  & 14.7  & 32.1 \\
    $\checkmark$      &       & $\checkmark$      &       &       & 54.2 & 40.7  & 31.2  & 17.9  & 35.7 \\
    $\checkmark$      & $\checkmark$      &       & $\checkmark$      &       & 54.7  & 40.2  & 32.8  & 20.6  & 36.2 \\
    $\checkmark$      &       & $\checkmark$      & $\checkmark$      &       & \textbf{56.2}  & \textbf{42.4} & 36.4  & 22.6  & 38.6 \\
    $\checkmark$      &       & $\checkmark$      & $\checkmark$      & $\checkmark$      & 55.4  & 42.0  & \textbf{39.2} & \textbf{24.5} & \textbf{40.6} \\
    \bottomrule
    \bottomrule
    \end{tabular}%
    }
  \label{teb6}%
  \vspace{-0.15in}
\end{table}%

\begin{figure}[!h]
  \centering
  \includegraphics[width=1.0\columnwidth]{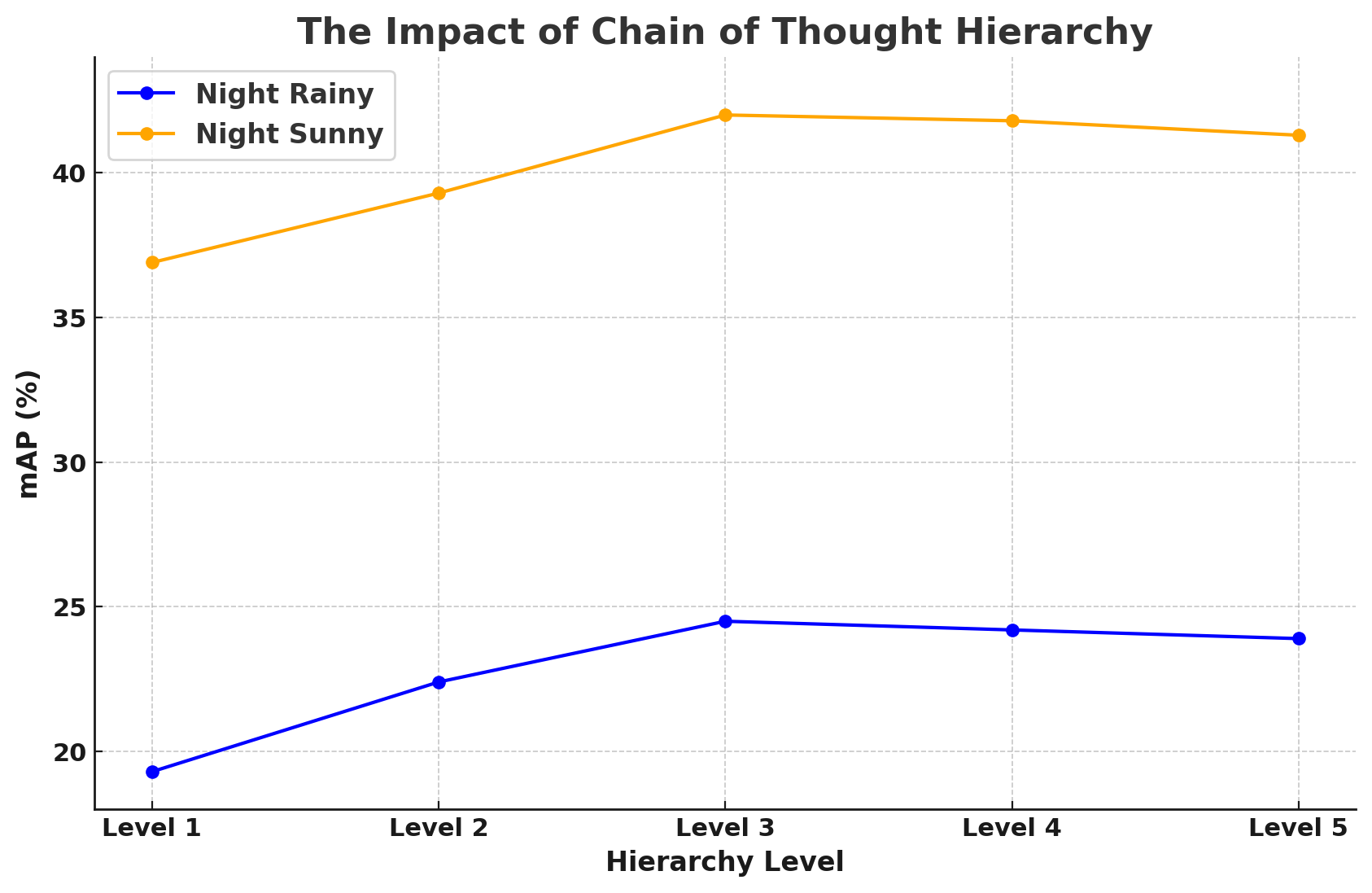}
  \vspace{-15pt}
  \caption{Analysis of Chain of Thought Hierarchy. Different levels represent the use of various the Chain of Thought Hierarchies.}
    \label{H}
    \vspace{-15pt}
\end{figure}

\textbf{The Impact of the Chain of Thought Hierarchy.}
Our style evolution guidance currently employs a three-tiered chain of thought, progressing from individual words to phrases, and finally to sentences, to gradually guide style development. Additionally, we evaluated the impact of different chain-of-thought levels on model performance, as shown in Figure \ref{H}. Level 1 represents the one-step prompt method, where all necessary information is provided in a single prompt, such as “Driving on a rainy night, heavy rain poured down, with some pedestrians and vehicles on the road.” Level 2 moves from words to phrases, simplifying the three-step prompt. Level 4 builds on Level 3 with added detail, while Level 5 introduces sentence templates, such as ‘a hard-to-see photo of a {}.’ or ‘a bright photo of a {}.’ However, likely due to limitations in current language model encoding capabilities, the three-tiered chain achieved the best results. With increasing levels, model performance decreased, potentially due to the influence of excessive information on style evolution.

\vspace{-5pt}
\subsection{Visualization Analysis}
\vspace{-5pt}
In Figure \ref{vis}, we present a comprehensive visual analysis of our method, primarily using heatmap visualizations to validate the effectiveness of style evolution guided by the chain of thought. The first column displays the model's detection results, directly reflecting its performance. The second column shows style transfer using a one-step prompt method. The third and fourth columns represent style evolution guided by two-level and three-level chain of thought, respectively. Meanwhile, we report the experimental performance of style evolution guided by different levels of the chain of thought in the supplementary material. Figure \ref{vis} demonstrates that as the chain of thought levels progresses, the model’s focus on the target significantly increases, with feature maps containing richer target information and reduced background interference. This confirms that gradual guidance through chains of thought enables the model to encounter a wider range of data distributions during training, enhancing its generalization to unseen target domains. 
\vspace{-5pt}

%% file: sec/5-Conclusion.tex
\vspace{-5pt}
\section{Conclusion}
For the Single-DGOD task, we propose a new method, i.e., Style Evolving along Chain-of-Thought, which continuously layers, integrates, and expands styles using text prompts that range from simple to complex and from coarse to fine-grained. This approach allows the model to encounter a richer variety of style features with different data distributions during training, thereby enhancing its generalization capability in unseen domains. Additionally, we introduce a Style Disentangled Module and Class-Specific Prototype to effectively perform style evolution. Experimental results show that our method outperforms others in detection across driving weather scenarios and Real-to-art benchmarks, validating its effectiveness.